\documentclass{article}

\usepackage{xspace}
\usepackage[inline,shortlabels]{enumitem}
\usepackage{amsmath}
\usepackage{booktabs}
\usepackage{hyperref}
\usepackage{multirow} 

\usepackage[numbers,sort&compress]{natbib}
\usepackage{todonotes}

\newcommand{\ppm}{Predictive Process Monitoring\xspace}
\newcommand{\PPM}{PPM\xspace}

\newcommand{\nirdi}{\texttt{Nirdizati}\xspace}

\newcommand{\hp} {hyperparameter\xspace}
\newcommand{\hps}{hyperparameters\xspace}

\newcommand{\eg}{\textit{e.g.}\xspace}

\newcommand{\cm}[1]{\textsc{#1}}
\newcommand{\fun}[1]{\textsl{#1}}
\newcommand{\cp}[1]{\ensuremath{\mathsf{#1}}}

\newcommand{\rdb}[1]{red box \textbf{#1}}

\pdfoutput=1

\begin{document}

\title{\nirdi: an Advanced \ppm Toolkit}


\author{Williams Rizzi$^{1,2}$, Chiara {Di Francescomarino}$^1$, \\
 Chiara Ghidini$^1$,  Fabrizio Maria Maggi$^2$\\
$^1$ Fondazione Bruno Kessler (FBK), Trento, Italy \\
$^2$ Free University of Bozen-Bolzano, Bolzano, Italy \\
{\{wrizzi, dfmchiara, ghidini\}@fbk.eu}, {maggi@inf.unibz.it}
}

\begin{abstract}
    Predictive Process Monitoring is a field of Process Mining that aims at predicting how an ongoing execution of a business process will develop in the future using past process executions recorded in event logs. The recent stream of publications in this field shows the need for tools able to support researchers and users in analyzing, comparing and selecting the techniques that are the most suitable for them. 
    \nirdi is a dedicated tool for supporting users in building, comparing, analyzing, and explaining predictive models that can then be used to perform predictions on the future of an ongoing case. By providing a rich set of different state-of-the-art approaches, \nirdi offers BPM researchers and practitioners a useful and flexible instrument for investigating and comparing Predictive Process Monitoring techniques. In this paper, we present the current version of \nirdi, together with its architecture which has been developed to improve its modularity and scalability. The features of \nirdi enrich its capability to support researchers and practitioners within the entire pipeline for constructing reliable \ppm models.
\end{abstract}

\maketitle


\section{Introduction}
\label{sec:introduction}

\ppm~(PPM)~\cite{DBLP:conf/caise/MaggiFDG14} is a branch of Process Mining that aims at providing predictions on the future of an ongoing process execution by leveraging past historical execution traces. An increasing number of PPM approaches leverage machine and deep learning techniques in order to learn from past historical execution traces the outcome of an ongoing process execution, the time remaining till the end of an ongoing execution, or the next activities that will be performed.

The importance of PPM has stimulated researchers and practitioners to create dozens of Machine/Deep Learning approaches to build PPM models \cite{DBLP:books/sp/22/FrancescomarinoG22}. With such a diverse range of Machine learning techniques an issue is how to sensibly choose the most appropriate one(s), especially considering that Machine Learning approaches performance can vary widely when the dataset supplied to build the PPM predictive model are different.

This paper proposes an architecture for a PPM toolkit - \nirdi - which offers researchers and practitioners with a rich set of features and techniques to \textbf{build, compare, and analyze} predictive process monitoring models. Given the growing importance of eXplainable AI (XAI), \nirdi is also equipped with a first set of features for explaining predictions. 
\nirdi has been implemented as an open-source toolkit. While stemming from the research community, \nirdi can be of interest to both practitioners and research scientists. Once compared and selected, the chosen predictive models can be used by practitioners to compute predictions on the future values of performance indicators related to a currently running process execution. Researchers instead can exploit the modular nature of the tool for extending it with new solutions, and leverage the standardized \nirdi pipeline to comparatively evaluate new solutions against existing techniques for better fitting models to data.

The structure of this paper is as follows.
Section~\ref{sec:background} provides a background on the data used in PPM, the techniques used to build and evaluate a predictive model, and the techniques used to explain the data and the predictive model to the user. Section~\ref{sec:capabilities} provides a description of the typical pipeline used for building predictive process monitoring models, and the capabilities a generic tool for PPM should satisfy. 
Sections~\ref{sec:architecture} and~\ref{sec:modules} describe the general \nirdi architectures and components to provide the capabilities described in Section~\ref{sec:capabilities}, respectively. Sections~\ref{sec:implementation} and~\ref{sec:use_case} provide an overview of the current prototype implementation and of some of its front-end features. The paper ends with an assessment of the tool (Section~\ref{sec:assesment}), related work (Section~\ref{sec:related_works}) and concluding remarks (Section~\ref{sec:conclusion}).  

\section{Background}
\label{sec:background}
In this section we provide an overview of the main concepts in Predictive Process Monitoring, and Machine Learning techniques and concepts used in the construction of the \nirdi tool. 

\subsection{Event Log}
\label{sec:background:data}

An \emph{event log} is composed of executions (also known as traces or cases) of business processes.
A \emph{trace} consists of a sequence of \emph{events}, each referring to the execution of an activity (a.k.a.~an event class). Events that belong to a trace are \emph{ordered} and constitute a single ``run'' of the process. For example, in trace $\sigma_i=\left\langle event_1, event_2, \ldots event_n \right\rangle$, the first activity to be executed is the activity associated to $event_1$.
Events are also characterized by attributes (\emph{event attributes}), among which the \emph{timestamp}, indicating the time in which the event has occurred, and possible \emph{data payloads} such as the resource(s) involved in the execution of an activity or other data recorded with the event. Some attributes, as for instance the personal birth date of a customer in a loan request process, do not change throughout the different events in the trace, and are called \emph{trace attributes}. 

The canonical representation format of the event logs is the XES (eXtensible Event Stream) standard~\cite{DBLP:conf/caise/VerbeekBDA10}: an XML-based format that maintains the general standard structure of an event log. 

\subsection{\ppm}
\label{sec:background:predictive_model}
\ppm~(\PPM)~\cite{DBLP:conf/caise/MaggiFDG14} is a branch of Process Mining that aims at predicting at runtime and as early as possible the future development of ongoing cases of a process given their uncompleted traces. In the last few years a wide literature about \ppm techniques has become available - see \cite{DBLP:conf/bpm/DiFrancescomarino18} for a survey - mostly based on Machine Learning techniques. 

\begin{figure*}[t]
  \centering
    \includegraphics[width=.7\textwidth]{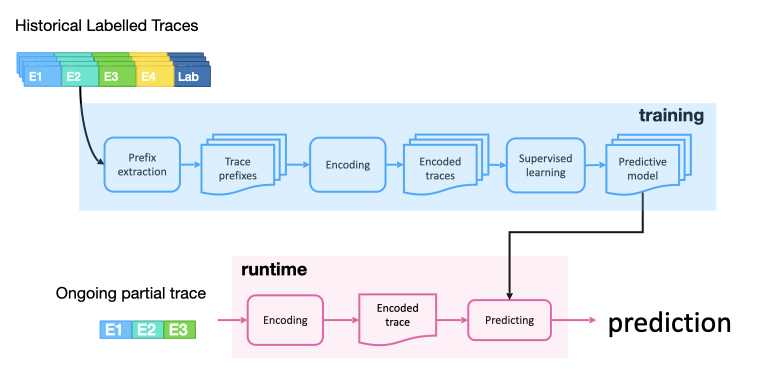}
  \caption{The \ppm pipeline.}
  \label{fig:images_PPM-pipeline}
\end{figure*}

Figure~\ref{fig:images_PPM-pipeline} shows a typical example of \ppm pipeline. These approaches usually require that trace prefixes are extracted from the historical execution traces (\emph{Prefix extraction phase}). This is due to the fact that at runtime predictions are made on incomplete traces, so that correlations between incomplete traces and what we want to predict (\emph{target variables} or \emph{labels}) have to be learned in the training phase. After prefixes have been extracted, prefix traces and labels (i.e., the information that has to be predicted) are encoded in the form of feature vectors (\emph{encoding phase}). Encoded traces are then passed to the (supervised learning) techniques in charge of learning from the encoded data one (or more) predictive model(s) (\emph{Learning phase}). At runtime, the incomplete execution traces i.e., the traces whose future is unknown, should also be encoded as feature vectors and used to query the predictive model(s) so as to get the prediction (\emph{Predicting phase}).

Concerning ``what we want to predict'', or better the label we aim at predicting, we can follow the typical classification presented in~\cite{DBLP:conf/bpm/DiFrancescomarino18} and cluster predictions into three main categories:
\begin{itemize}
\item \emph{outcome-based predictions}, that is, predictions related to predefined categorical or boolean outcome values; 
\item \emph{numeric value predictions}, that is, predictions related to measures of interest taking numeric or continuous values; and 
\item \emph{next event predictions}, that is, predictions related to sequences of future activities and related data payloads.  
\end{itemize}

Encodings are needed to transform labelled prefix traces of the form $\sigma_i=\left\langle event_1, event_2, \ldots event_i, label \right\rangle$ into a format that is understandable by machine learning techniques. This would allow the technique to train and hence learn, from encoded data a predictive model. Due to the complex nature of event logs, which are characterized by a temporal dimension, the presence of data payloads, the possible inter-relation vs independence between execution traces, and so on, several encodings have been proposed in the literature. 
Typical encodings available in the literature~\cite{DBLP:conf/bpm/LeontjevaCFDM15} are: the \emph{Boolean encoding} and the \emph{Simple-index encoding}, which only take int account the control flow (that is, the sequence of events) of a trace, and the \emph{Complex index-based encoding}, which instead also consider data payloads in the encoding. Further encodings, such as the inter-feature encoding presented in \cite{DBLP:conf/bpm/SenderovichFGJM17}, make ti possible to machine leaning algorithms to take into account the possible interdependencies between temporally overlapping execution traces as this may affect the result of a prediction (think, for instance, to ``runs'' competing for the same resource).  

When it comes to the learning phase, multiple approaches from the machine learning literature have been used in the literature of \ppm. Examples range from 
traditional Machine Learning techniques such as Decision Tree, Random Forest, or Support Vector Machine to deep learning techniques such as LSTM Neural Networks~\cite{DBLP:journals/is/LeoniAD16,DBLP:conf/caise/MaggiFDG14,DBLP:journals/tsc/Francescomarino19,DBLP:conf/bpm/LeontjevaCFDM15,DBLP:conf/caise/TaxVRD17}. 
One of the reasons for such a variety of techniques is that the studies carried out have not identified a Machine Learning algorithm strictly better than all the others. Depending on the event log data, the encoding, and the prediction problem, different techniques are known to perform better than others. 

\subsection{Hyperparameters optimization}
The machine learning techniques to better fit the supplied data typically have a variable number of \hps. The tuning of the \hps highly influences the quality of the built predictive model. 
Indeed, although the available techniques are expensive both in terms of computational power and time, the gains in terms of prediction quality justify their use. 

The \hp optimization techniques receive as input the \hps space of correct input and a quality metric to maximize. The technique will then proceed to maximizing the quality metric by producing multiple different configurations of \hps and then evaluating the quality metric for each of them. The amount of \hp configurations produced by the \hp optimization technique depends on how the technique explores the \hp optimization space. Various algorithms exist to tackle this challenge~\cite{DBLP:conf/nips/BergstraBBK11}, including approaches specific for the \ppm field~\cite{DBLP:journals/is/Francescomarino18}.


\subsection{Explainable Predictions}
\label{sec:background:explanation}
Explainable AI (XAI) is a growing body of research focused on providing explanations on the outputs of, usually gray or black box, Machine Learning algorithms. In the literature, there are two main sets of techniques used to develop explainable systems, a.k.a.\ explainers: \emph{post-hoc} and \emph{ante-hoc} techniques. Post-hoc techniques, such as the Local Interpretable Model-Agnostic Explanations (LIME) \cite{DBLP:journals/corr/RibeiroSG16} or the SHapley Additive exPlanations (SHAP) \cite{DBLP:conf/nips/LundbergL17}, allow models to be trained as usual, with explainability only being incorporated at testing time. Ante-hoc techniques, such as Bayesian deep learning (BDL) \cite{DBLP:journals/compsys/DenkerSWSHJH87}, entail integrating explainability into a model from the beginning.
Explainable techniques have recently started to be applied to the \ppm field, mainly with the usage of model agnostic generic techniques (such LIME and SHAP mentioned above) with the aim of discovering the rationale for a prediction. While the application of XAI techniques to PPM is still at its infancy, and specific XAI techniques devoted to event log data or \ppm prediction problems are still lacking, the field is developing quickly. The interested reader is referred to \citep{DBLP:conf/ecis/StierleBWZM021} for a recent survey.

\section{Building a \ppm model}
\label{sec:capabilities}
\begin{figure*}[t]
  \centering
    \includegraphics[width=.8\textwidth]{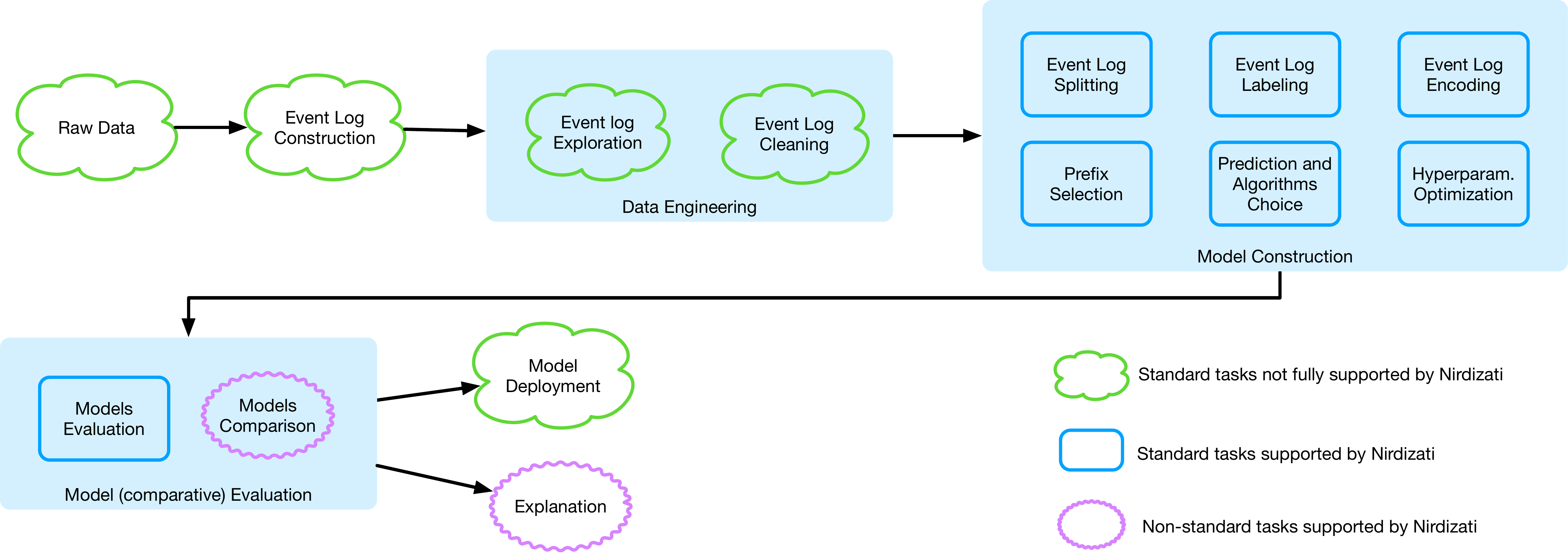}
  \caption{Building a \ppm model.}
  \label{fig:images_pipeline}
\end{figure*}

The task of building a \ppm model can be seen as an instance of a data science pipeline, tailored to the specific type of event logs data. While several data science pipelines exist, they are all usually composed of a set of tasks  
that start with the raw data, transform them into event logs, and then move trough the steps of Data Engineering, Model Construction (and Evaluation), and Model Deployment (see \figurename~\ref{fig:images_pipeline}). 
	
\nirdi aims at supporting users in the \emph{construction and the comparison of different predictive models} so as to: 
\begin{enumerate*}[(i)]
\item select the most appropriate model(s), in case of end users who aim at deploying a predictive model, or 
\item support experimental comparisons of different techniques in scientific studies, in case of research scientists.
\end{enumerate*}
As such, it mainly focuses on providing capabilities for Model Construction and its (comparative) Evaluation, as highlighted in the blue boxes in \figurename~\ref{fig:images_pipeline}, and neglects steps such as the extraction of raw data, the construction of an event log, and the deployment of the predictive model, which are often tackled with ad hoc approaches or different specific tools (see \eg the \emph{onprom} tool for the construction of an event log, described in \cite{DBLP:conf/bis/CalvaneseKMT17}). Also, the Data Engineering component is mainly tailored to support the splitting and labeling of data into training and test sets, thus ignoring the important - but often similarly ad hoc - step of Data Cleaning and Improving. Given the growing importance of Explainable Artificial Intelligence (XAI), explanation is likely to become one of the standard tasks of future ML-based pipelines and is therefore also included in \nirdi and discussed here.   
In the following we provide an overview of the capabilities (that is the functionalities for the user) that a tool for constructing and comparing   different \ppm models should have, ranging from the creation of a good labeled training and test set, up to the construction of \ppm models, and their (comparative) evaluation.

\subsection{Data Engineering}
\label{sec:capabilities:data}
In the context of \ppm, data pre-processing is used to determine the most effective representation of the data for the predictive problem at hand. In particular, data pre-processing is used in \ppm to understand the distribution of attributes, events, and timestamps in the data space, and whether the quality of data and their interesting aspects make the specific prediction problem at hand feasible, unfeasible, or only partly feasible. 

The understanding of the quality of data and the feasibility of some/all of the canonical prediction problems is usually tackled with data exploration, which is concerned with the determination of the data types, the visualisation of such data, and the analysis of the data space densities.
In the context of \ppm data exploration must enable the user with the \textbf{exploration of the selected event log}, which in turn consists in the exploration of execution traces in the XES format, e.g., their start and end time, the distribution of events, as well as of event or trace attributes. 
Data exploration can trigger a refinement of the data space, which is usually performed in a data cleaning phase, which can concern the removal of unnecessary,  duplicate, or inconsistent values, the addition of missing values, or the modification of the sparsity of data, just to make simple examples. 
Several open source or commercial tools exist to perform complex event log exploration, analysis, and engineering tasks, usually as a prerequisite of the task of process discovery (see e.g., ProM\footnote{\url{https://www.promtools.org}}, Disco\footnote{\url{https://fluxicon.com}}, Celonis\footnote{\url{https://www.celonis.com/}}, just to name a few). Thus, supporting advanced event log engineering functionalities is not deemed crucial for a \ppm tool. Obviously an advanced tool for PPM \emph{may enable} the user to perform some event log engineering as a way to support the user within a single comprehensive tool. 

\subsection{Building the predictive model(s)}
\label{sec:capabilities:predictive_model}

Once the event logs are in a suitable shape, they must be \textbf{split} into training (the data used to build the predictive models), validation (the data used to automatically optimize hyper-parameters, if used) and test (the data used to evaluate the predictive models) sets. From here on, we will use the term \emph{training set} to denote the data used for building a model with optimised hyperparameters, that is for denoting both the training and validation set.   
In addition to the choice of the size of training and test sets, the temporal nature of \ppm data implies that different strategies can be chosen in order to split them.  For instance one may opt for a random split of traces for training and testing disregarding the time they were executed; one could take the older\footnote{That is, the ones starting (resp., finishing) at the beginning of the period covered by the event log}  for training and the newer\footnote{That is, the ones starting (resp., finishing) towards the end of the period covered by the event log} for testing so as to implement a ``not learn from the future'' strategy; or one may decide further preferred ``orderings''. 
As a consequence, \emph{an advanced tool for PPM should enable the user to split an event log into training and test sets by choosing an appropriate split strategy}. 

Another important step in preparing the training and test set is the choice of ``when to predict'' in the ongoing trace, that is the length(s) of the sequences (prefixes) to be used to train the predictive model. Similarly to the previous step, several strategies exist in \ppm to \textbf{select prefixes} in the construction of a predictive model, which need to be supported by an advanced \ppm tool. For example, one can choose to build a model for predicting exactly at a prefix of length $n$ (that is, after $n$ events in a sequence), for predicting at any prefix of length up to $n$, or to build $n$ different predictive models, one for each prefix of length $i \leq n$. A different strategy would be to choose to predict after (the first occurrence of) a given activity ``A'' regardless of when it happens in the trace. 
Once the prefix is chosen, an advanced \ppm tool must enable the user to decide how to deal with traces that are shorter than the selected prefix length. Again, different strategies are possible: for example, sequences shorter than $n$ can be discarded, or they can be padded with ``0'' values by adopting a standard strategy when dealing with input sequences of variable length. 
As a consequence, \emph{an advanced tool for PPM should enable the user to select the prefixes used to build the predictive model and the appropriate prefix selection strategy}. 

A similarly important step in building a training and test set is the support for the \textbf{labeling} of execution traces. While providing support for any kind of automatic labeling is rather challenging, an advanced tool for PPM should at least provide support for typical types of labels. Examples are: attributes used for outcome prediction, next activity(ies), categorical values referring to the duration of the trace (e.g. fast or slow w.r.t. a given value or some average trace duration), and so on. Thus,     
\emph{an advanced tool for PPM should provide extensive support for automatic labeling of training and test data. }

Execution traces are not in a format that can be directly used by a Machine Learning algorithm. As shown in Section~\ref{sec:background:predictive_model} different \textbf{encodings} have been proposed in the literature \cite{DBLP:conf/bpm/LeontjevaCFDM15} so as to exploit different characteristics of the event logs. Therefore, a key capability of an advanced \ppm tool is the availability of different encoding methods for the dataset at hand. Furthermore, the \ppm literature has shown that \ppm approaches can predict an \eg outcome of a running case as dependent (resp. independent) on the interplay of all cases that are being executed concurrently \cite{DBLP:conf/bpm/SenderovichFGJM17}. Together with the choice of the encoding method, an advanced \ppm tool must enable the user to specify whether traces are considered in an \textbf{intercase vs. intracase fashion}.  
Thus, \emph{an advanced tool for PPM should enable the user to select among different encodings} and \emph{whether to perform predictions in an intercase or intracase fashion}.

When it comes to the \textbf{selection of the Machine Learning (ML) algorithm} to be used for training a model, it is well known that ``\emph{no silver bullet exists}''. Indeed, while it is true that some findings exist~\cite{Teinemaa2019} one of the challenges of the \ppm field is exactly to understand which specific algorithm is better suited for specific data or predictive problems.
To tackle the issue of building the most appropriate model for the given data and prediction problem, an advanced \ppm tool should provide support for a variety of techniques together with their parameter tuning.  
Thus, \emph{an advanced tool for PPM should enable the user to select among a wide variety of techniques and their configurations}. 

Moreover, and most importantly, an advanced tool for PPM should enable the user to \textbf{compare configurations} emerging from all the different choices illustrated above. That is, 
\emph{an advanced tool for PPM should enable the setting up of different configurations covering all the steps above (from splitting, to the selection of the ML algorithms and their parameters) and the comparative construction of all their corresponding predictive models}. 

The ability to store a number of alternatives - from the encodings to the ML techniques - and the need to run the (often expensive) computation of several predictive models at once imposes requirements on the software architecture that will be discussed in Section~\ref{sec:architecture}. For what concerns the tool support towards model construction, the building of several models triggers the need to support some kind of \emph{hyperparameter optimization}. Indeed the optimization of the hyperparameters of a ML algorithm is known to be a difficult and time expensive task, and end users may find unfeasible to autonomously optimize the hyperparameters of multiple ML algorithms. Thus \emph{an advanced tool for PPM should enable the computation of an optimal configuration for each of the supported ML algorithms}.


\begin{table*}[tb]
	\centering
\resizebox{\columnwidth}{!}{
	\begin{tabular}{@{}ll@{}} 
	\toprule     
        (\cp{C2}) & support the split event log into training and test sets with appropriate split strategy\\
        (\cp{C3}) & support the selection of prefixes and the appropriate prefix selection strategy\\
        (\cp{C4}) & support for automatic labeling of event logs\\
        (\cp{C5}) & support for the selection of the appropriate encoding of event logs\\
        (\cp{C6}) & support intercase and intracase predictions\\
        (\cp{C7}) & support the choice among a variety of configurable ML techniques\\
        (\cp{C8}) & support the construction of different models \\
	(\cp{C9}) & support automatic hyperparameter optimization \\
	(\cp{C10}) & support different evaluation metrics, including the accuracy ones\\
	(\cp{C11}) & support models comparison in terms of metrics and characteristics of the configuration\\
	\midrule
	(\cp{C1}) & support event log exploration and engineering\\
	(\cp{C12}) & support explanations of predictions\\
	\bottomrule
	\end{tabular}
}
	\caption{The capabilities of a \ppm tool.}
	\label{table:capabilities}
\end{table*}

\subsection{Evaluating the predictive model(s)}

The evaluation step is of tantamount importance, when aiming at the construction of a predictive model that is appropriate for the dataset at hand. A tool must here support the user in two dimensions: the possibility to rely on different metrics for each model, and the capability to comparatively evaluate all the different configurations defined to build different models.  

Focusing on the \textbf{metrics} to evaluate each single model, it is well known in Machine Learning that different evaluation metrics are used for different kinds of problems. Thus, \emph{an advanced tool for PPM should enable the choice between a wide set of evaluation metrics}. In particular, it should enable the choice of appropriate accuracy metrics, as these metrics provide an estimate of how well the predictive model understood the supplied data and is capable to exploit new data to compute a value satisfying the prediction problem. 
In addition, it should be able to accommodate further metrics useful to scientists in their experimental evaluations of \ppm techniques. An example of such metrics are time metrics. These metrics measure the amount of time spent by the tool to perform a set of actions, such as e.g., the time spent in pre-processing the data, or the time spent in building the predictive model. 

Focusing on the \textbf{comparative evaluation}, the comparison between different predictive models needs to be supported in terms of (at least) two different instruments. The first instrument is composed of visual (e.g, tabular) representations enabling the comparison of the different metrics illustrated above. In that, a user may investigate which configurations have better, say, F1 or AUC scores. This is a basic evaluation capability which could help for instance an end user directly selecting the best performing model. The second instrument should enable the user to drill down and understand how the goodness of the model (e.g. the AUC) is related to a certain configuration step, for example the prefix length, the encoding, or the chosen ML technique. This latter more sophisticated method may be useful to 
\begin{enumerate*}[(i)]
	\item tailor future configurations for the construction of further models, in case of not fully satisfactory ones, and   
	\item carry out scientific investigations, where it is often required to understand why performances may vary.  
\end{enumerate*} 
Thus, \emph{an advanced tool for PPM should enable the comparison of different predictive models both in terms of (performance) metrics and in terms of characteristics of the configuration used for model construction}.

\subsection{Explaining the predictive model(s) behaviors}

Given the growing importance of Explainable Artificial Intelligence (XAI), explanation is likely to become one of the standard tasks of future ML based pipelines. Therefore, even if this task is not yet part of a standard \ppm pipeline, a good \ppm tool should include the possibility of easily \textbf{incorporating XAI techniques}, both to explain a particular prediction or model and to be added as one of the criteria upon which models can be compared. 
In addition to computing the explanation, a \ppm tool should be apt to tailor and display the explanation to the target user. 
Given the recent nature of the XAI field, and the different types of ML techniques, data, and prediction problems, no general studies exist that indicate which XAI techniques are the best ones for what.
Therefore \emph{an advanced tool for PPM should enable the usage of different XAI algorithms and of different visualization methods.}

Table~\ref{table:capabilities} summarizes all the capabilities of a \ppm tool illustrated in this section. As discussed in the section, capabilities (\cp{C2})--(\cp{C11}) are considered necessary functionalities of an advanced tool for PPM, while capabilities (\cp{C1}) and (\cp{C12}) are considered useful extra features part of which - as in the case of (\cp{C12}) - may become necessary in the future. In the next section, we describe the architecture of \nirdi which is tailored towards supporting the above-mentioned capabilities.

\begin{figure*}
    \centering
    \includegraphics[width=.7\textwidth]{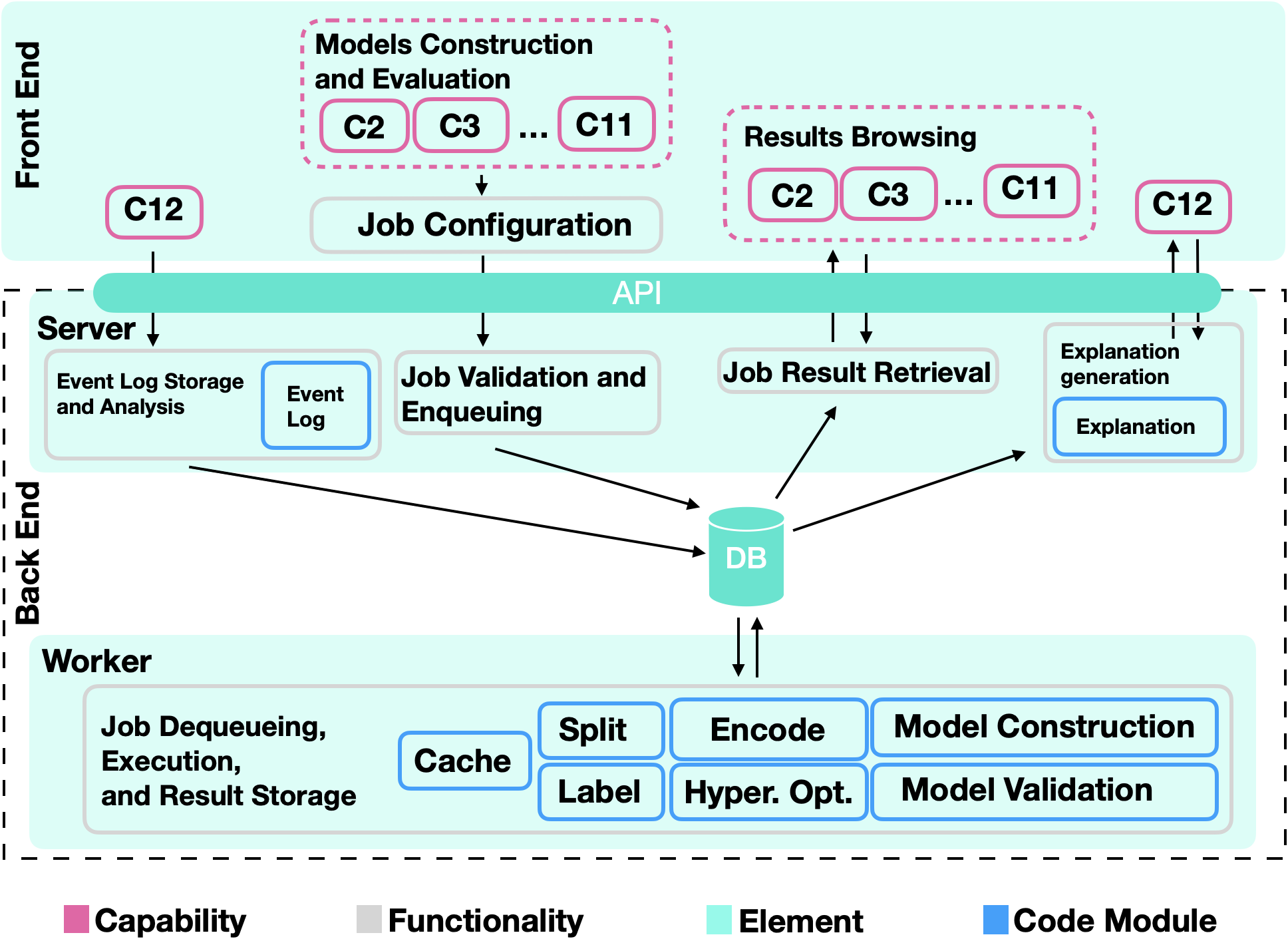}
    \caption{The capabilities of \nirdi are provided by the concertation of multiple architectural elements. The direction of the arrows represents the flow of information. The user accesses the capability interacting with the Front-end element. In turn, the Front-end element sends requests to the Server element via the APIs. The Server, the Database and the Workers provide the functionalities: the 
	simple requests are handled by the Server or the Server and the Database, whereas the Job requests are handled by the Worker.}
    \label{fig:architecture:choreography}
\end{figure*}

\section{The \nirdi Architecture}
\label{sec:architecture}

The support of complex capabilities as those mentioned in Section~\ref{sec:capabilities} requires both the ability to offer friendly user interfaces and the ability to ensure fast processing and a good usage of resources, e.g., to train the models, so as to exploit the available computational capabilities in an efficient manner.

\begin{figure*}
    \centering
    \includegraphics[width=.6\textwidth]{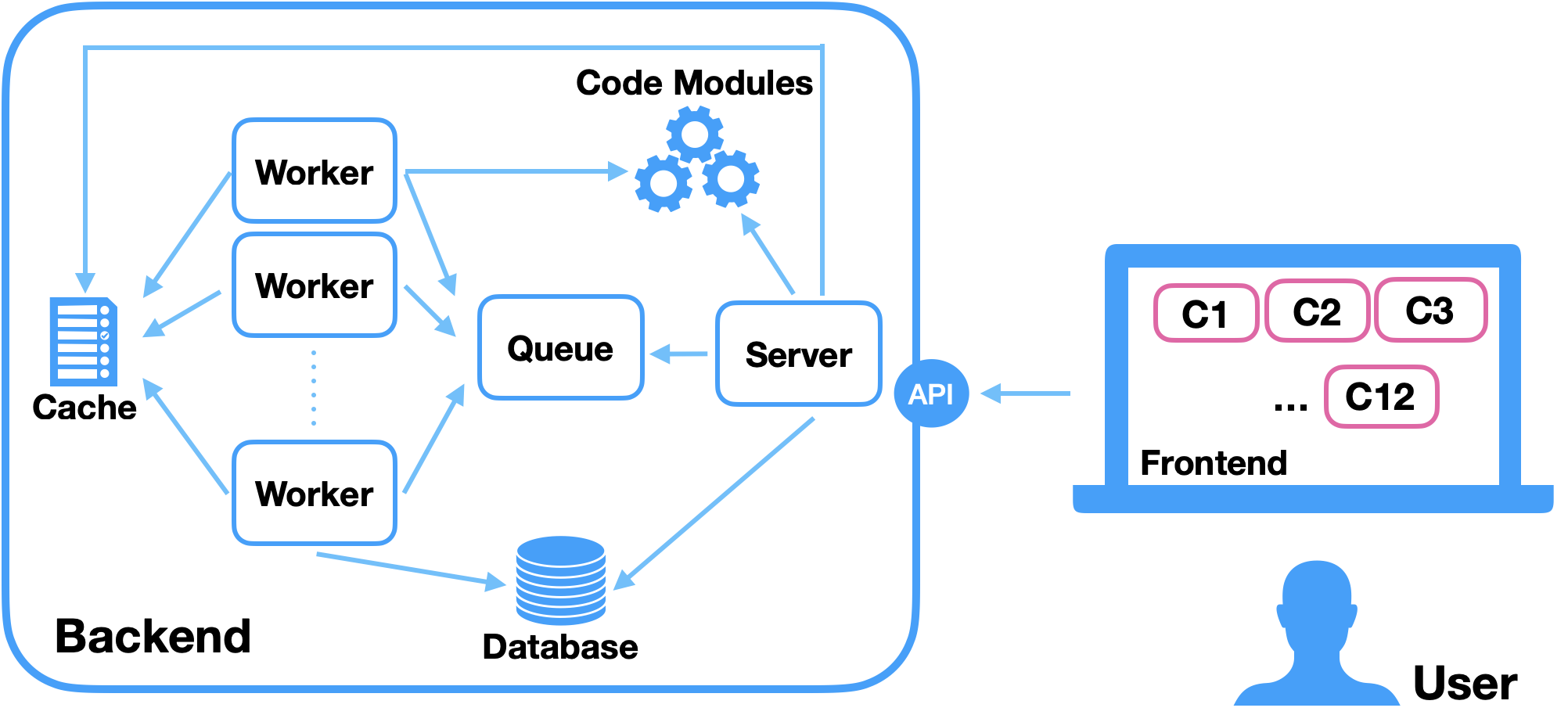}
    \caption{\nirdi follows a client-server architecture. The server exposes APIs to communicate with the client. When the server receives a new job, it puts it in a Queue. When a new job is available in the Queue, a worker takes care of the job. Workers, whose number can vary according to the needs, share access to the Queue, to the Cache, and to the Database.}
    \label{fig:architecture:master-slave}
\end{figure*}

At the high level, \nirdi is composed of (i) a Front-end application, which allows users to select the prediction methods and to assess the goodness-of-fit of the built models, by means of a full / partial implementation of capabilities (\cp{C1}) -- (\cp{C12}), and (ii) of a Back-end application responsible for the actual
training and test (see \figurename~\ref{fig:architecture:choreography}). Our aim to offer \nirdi as a service to the research community has triggered the decision to built it as a web application. We postpone the description of the different code modules, and their relations with  capabilities (\cp{C1})--(\cp{C12}), to Section~\ref{sec:modules}. Instead we focus here on the architectural design that we adopted in order to support modularity and scalability of the processing (see \figurename~\ref{fig:architecture:master-slave}). 

In detail, \nirdi\footnote{Source code available at \url{https://github.com/nirdizati-research/}} supports the multiprocessing capabilities of nowadays processors, using a Master/Replica setting. 
In this setting, the Master - the \emph{server} in \figurename~\ref{fig:architecture:master-slave} - creates and orchestrates the jobs and performs other small tasks, thus providing functionalities such as the \fun{Job Validation and Enqueueing}, the \fun{Job Result retrieval}, the \fun{Event Log Storage and Analysis} and the \fun{Explanation Generation}. The Replica - the \emph{workers} in \figurename~\ref{fig:architecture:master-slave} - instead, take care of performing (most of) the jobs (\fun{Job Dequeueing, Execution and Result storage}). 
They do it by exploiting \emph{code modules} that implement the required functionalities in a modular setting (see Section~\ref{sec:modules}).

When using such a setting for comparing different algorithms on the same task, most of the processing time is taken for the pre-processing phases that are repeated over and over for each different configuration.
To overcome this time and resource consumption, \nirdi~has been equipped with a \emph{cache system} that avoids pre-processing the data multiple times when evaluating multiple approaches on a shared task and dataset.
When using caching in the \nirdi~architecture, however, two or more slaves performing the same pre-processing step concurrently can generate duplicate cache files.
A \emph{database system} able to track in a granular way the status of each object of the tool, i.e., the set of elements used in a given task, has hence been introduced.
For instance, the object representing the task of splitting a log into a training and a test set is composed of a foreign key to the object representing the log, the training and the test set sizes, the type of ordering used for the split, the name of the split, and a unique key.
Storing each object of the tool in a table of the database allows, on the one hand, an easy retrieval of the required information and, on the other hand, the capability to keep in memory only the database keys instead of the objects themselves, thus resulting in a lighter footprint in the memory.
From the database it is possible to retrieve the current jobs and their status, the available cached information, the previously trained models and the corresponding accuracy metrics.
By relying on a stateful architecture and capitalizing the work carried out, \nirdi~is able to exploit multiprocessing.

\begin{table}[t]
    \centering
        \begin{tabular}{ l c c c }
	\toprule	
                     & \multicolumn{3}{c}{Capabilities} \\
	\cmidrule(rl){2-4}	
        Element      & [\cp{C1}, \cp{C7}, \cp{C11}] & [\cp{C3}, \cp{C12}] & [\cp{C2}, \cp{C4}--\cp{C6}, \cp{C8}--\cp{C10}]  \\
        \midrule
        Front-end     &      x      &               x             &    x    \\
        API	     & 	    x      &               x             &    x    \\
	Server       &      x      &               x             &    x    \\
        Database     &      x      &               x             &    x    \\
	Code         &      x      &               x             &    x    \\
        Queue        &             &                             &    x    \\
        Cache        &             &               x             &    x    \\
        Worker       &             &                             &    x    \\
        \bottomrule
        \end{tabular}
    \caption{Architectural Elements and Capabilities.}
    \label{table:elements}
\end{table}

Table~\ref{table:elements} describes how the different architectural elements of \figurename~\ref{fig:architecture:master-slave} are involved in the provision of Capabilities (\cp{C1})--(\cp{C12}). For the sake of readability we have grouped capabilities that involve the same architectural elements. The presence of a ``x'' on a given cell means that the Architectural Element of that row is used to provide the group of Capabilities of that column. For instance, the provision of \cp{C1}, \cp{C7}, and \cp{C11} involves the Front-end, the API, the Server, the Database, and some Code Module.

Not surprisingly, the Front-end, the API, the Server, the Database, and Code Modules are involved in the provision of all capabilities. The first two groups of capabilities do not require any intensive computation\footnote{Actually this may change for the provision of explanations (\cp{C12}), which are currently handled by the Server only because of the still investigative nature of this functionality.} and are therefore handled directly by the Server, with the only difference of (\cp{C3}) and (\cp{C12}) using the Cache for optimization purposes. The third group, instead, is the one involved in the setting up, computation, and comparison of the different predictive models. Being time and resource expensive, the provision of these capabilities takes advantage of the \nirdi Master/Replica architecture based on the usage of Workers and Caching.

\section{Code Modules}
\label{sec:modules}
To ensure modularity, capabilities (\cp{C1})-(\cp{C12}) are supported in \nirdi by means of nine specific code modules: 
\cm{Event Log},
\cm{Split},
\cm{Label},
\cm{Encode},
\cm{Model Construction},
\cm{Cache},
\cm{Hyperparameter Optimization},
\cm{Model Evaluation}, and
\cm{Explanation}, as summarized in Table~\ref{table:modules}. Each module provides simple functionalities that compose the internals of the elements discussed in Section~\ref{sec:architecture} and allows for the provision of the complex capabilities discussed in Section~\ref{sec:capabilities}. In the following we briefly describe the different modules by highlighting their main characteristics and thus providing an illustration of how they support the required capabilities.

\begin{table}[tb]
    \centering
    \resizebox{\columnwidth}{!}{%
        \begin{tabular}{ l c c c c c c c c c c cc }
        \toprule
                                    & \multicolumn{12}{c}{Capabilities} \\
				    \cmidrule(rl){2-13}	
        Modules                     & \cp{C1}& \cp{C2} & \cp{C3} & \cp{C4} & \cp{C5} & \cp{C6} & \cp{C7} & \cp{C8} & \cp{C9} & \cp{C10}& \cp{C11}& \cp{C12}\\
        \midrule
        \cm{Event Log}                   & x & x &   &   &   &   &   &   &   &   &   & x \\
        \cm{Split}                       &   & x & x & x & x & x &   &   &   &   & x &   \\
        \cm{Label}                   &   &   &   & x &   &   &   &   &   & x & x &   \\
        \cm{Encode}                    &   &   &   &   & x & x &   &   &   &   & x &   \\
        \cm{Model Construction}           &   &   &   &   &   &   & x & x &   & x & x & x \\
        \cm{Hyper. Opt.} &   &   &   &   &   &   &   &   & x &   &   &   \\
        \cm{Model Evaluation}              &   &   &   &   &   &   &   &   &   & x & x &   \\
        \cm{Explanation}                 &   &   &   &   &   &   &   &   &   &   &   & x \\
        \cm{Cache}                       & x & x & x & x & x & x &   & x &   & x &   & x \\
        \bottomrule
        \end{tabular}
    }
    \caption{Code Modules and Capabilities.}
    \label{table:modules}
\end{table}

\paragraph{\cm{Event Log}}
\label{sec:modules:event_log}
The \cm{Event Log} module receives as input the data uploaded by the user, in the form of an event log in the XES Format, and stores them in the Back-end ready to be used. In order to help the user to establish which predictive model is most likely to achieve better performance~\cite{DBLP:conf/bpm/DrodtWMD21}, a description of the event log in terms of level of contention of resources in the data, parallelism in the execution, and sparsity of the data is computed and stored at upload time.

\paragraph{\cm{Split}}
\label{sec:modules:event_log:split}
The \cm{Split} module receives as input an event log and deals with the production of the training and the test sets. 
These sets are subsets of the initial event log (training and test logs), and their definition influences the quality of the trained predictive models. To produce the training and test sets, the \cm{Split} module needs to support the sorting, filtering, and splitting of the event log. 
The sorting of the event log allows for taking into consideration different perspectives of the data, such as the seasonality or the variants representation. 
The filtering of the event log allows for eliciting and correcting underrepresented variants, wrongly represented traces, malformed traces, biased traces, and in general information that we are not interested in representing.
The split of the event log allows for obtaining the training and test sets. The relative sizes of the sets highly influence the final performance of the predictive model; the split of the event log, indeed, might create truncated traces, overlaps between the sets, or skewed sets, that can invalidate the predictive model evaluation, as discussed in~\cite{DBLP:conf/bpm/WeytjensW21a}.
In addition to the above, the \cm{Split} module also supports prefix selection. 

\paragraph{\cm{Label}}
\label{sec:modules:event_log:labeling} 
The \cm{Label} module receives a training or test event log and returns the same log annotated with the desired label, that is, each trace in the event log is annotated with the label of interest. 
Typical labels in \PPM refer to categorical outcomes, numeric values, and next activities. The \cm{Label} module of \nirdi enables the automatic labeling of a log with all three types of predictions. 
In detail, for categorical outcomes, the module allows for (i) multiclass labels starting from categorical attributes and from the next activity, as well as (ii) binary labels starting from  numerical attributes and trace durations by leveraging a threshold value. For numeric outcomes, the module allows for numeric labels by starting from numeric attributes and trace duration.  

\paragraph{\cm{Encode}}
\label{sec:modules:event_log:encoding}
The \cm{Encode} module receives a labeled training/test log, and returns its encoded version as a Dataframe\footnote{\url{https://www.databricks.com/glossary/what-are-dataframes}}. To produce the Dataframe from a log, three different steps need to be tackled:
\begin{enumerate*}[label=(\roman*)]
		\item encoding information extraction, 
    \item feature encoding, and 
    \item data encoding.
\end{enumerate*}

The \emph{encoding information extraction} step receives as input a training/test event log and returns some useful information for the encoding. In this step the attribute(s) of the event log related to the control-flow (i.e., the attribute holding the activity value), to the data flow (i.e., trace and event attributes), as well as to the resource-flow (i.e., the attribute related to the resource) are extracted. This mapping is useful in order to understand what kind of information to take in (more) account in the actual encoding.

The \emph{feature encoding} step receives as input a training/test log and the encoding information and returns the set of features that will be used to represent each trace in the training/test sets. This step takes also care of encoding in a way to support intercase or intracase predictions. 

The \emph{data encoding} step receives as input the training/test log and the feature set and returns the training/test Dataframe and the encoder object. The Dataframe is built to be easy to understand for a machine and allows for a smoother training. The operations performed to build a machine-understandable object include steps like the one-hot encoding of categorical features, de-noise and normalization for the numeric object categorization of numeric features.

\paragraph{\cm{Model Construction}}
\label{sec:modules:predictive_model}
The \cm{Model Construction} module receives the predictive model specification and returns an instantiation of the predictive model with the specification of the supported hyperparameters space. The \cm{Model Construction} module is concerned with the correct initialization of the predictive model. There are three classes of predictive model building algorithms supported by \nirdi namely, (i) classification, (ii) regression, and (iii) clustering. These three classes 
 allow us tackling the three classic types of predictions: (i) categorical outcomes, (ii) numerical values, and (iii) next-activities, respectively. In turn, each class of predictive model building algorithms includes different specific algorithms that can be further
 increased, due to the modular nature of the tool. Section~\ref{sec:use_case:train} provides a description of the Front-end features of this module.

\paragraph{\cm{Hyperparameter Optimization}}
\label{sec:modules:predictive_model:hyperparameter_optimization}
The \cm{Hyperparameter Optimization} module receives in input the training 
Dataframe, an instantiated predictive model, and the predictive model hyperparameter space, and returns the optimized predictive model. Machine Learning techniques are known to use model parameters and hyperparameters. While the values of hyperparameters can influence the performance of the predictive models in a relevant manner, their optimal values highly depend on the specific dataset under examination, thus making their setting rather burdensome.
To support and automatize this onerous but important task, several hyperparameter optimization techniques have been developed in the literature~\cite{Bergstra-Hyperopt} and included in the \cm{Hyperparameter Optimization} module.

\paragraph{\cm{Model Evaluation}}
\label{sec:modules:evaluation}
The \cm{Model Evalauation} module receives in input the test Dataframe and the (hyper)optimized predictive model and returns the evaluation of the model using the selected metrics.
There are two classes of metrics that one can use in \nirdi: (i) time and (ii) accuracy metrics. Time metrics are concerned with understanding how fast the predictive model can train, update and return predictions. 
Accuracy metrics, instead, are concerned with understanding how well the predictions are in comparison with the ``perfect predictions'' contained in the test set. Being one of the distinctive features of \nirdi, the \cm{Model Evaluation} module provides a wide range of information for the comparative evaluation of different predictive models based on their configurations, results and prefixes. In addition to comparative tables, it also supports the display of bubble chart visualizations. Section~\ref{sec:use_case:validate} provides a description of the Front-end features of this module.

\paragraph{\cm{Explanation}}
\label{sec:modules:visualisation:explanation}
The \cm{Explanation} module receives in input a Dataframe and the trained predictive model and returns one of the three explanation classes discussed in~\cite{rizzi2022explainable}.
The three classes of explanation refer to three different levels of abstraction an explanation in \nirdi can refer to: (i) event, (ii) trace, and (iii) event log.
At the \emph{event} level, the impact of each feature (and corresponding value) on the prediction returned by the predictive model at the current prefix length is shown to the user. At the \emph{trace} level, how the impact (in terms of SHAP values) of each feature (and corresponding value) on the prediction evolves at different prefix lengths is returned to the user. Finally, at \emph{log} level, the average value of the prediction over all the traces of the event log for each value of a given feature is shown to the user (together with the number of traces characterized by that value for the feature).
Section~\ref{sec:use_case:explain} provides an example of 
one of the features supported by these module.

\paragraph{\cm{Cache}}
\label{sec:modules:predictive_model:cache}
The \cm{Cache} module receives in input an object, such as an event log, a Dataframe, or a trained predictive model and stores it. The cache module determines an efficient storage position of the resource, proceeds to store the object and it retrieves it when needed. In other words, the \cm{Cache} module is concerned with the efficient administration of the loaded and created objects. The objects that are more suitable to be cached are the ones that take a long time to be computed and occupy a small amount of space on the disk. We hence cache the loaded log, the labeled dataframe, and the trained predictive model. The loaded log is an event log that is loaded starting from the XES format into an in-memory representation, that is then used for the log encoding. The labeled dataframe is the log that is already transformed in the Dataframe format. The trained predictive model is the predictive model that is ready to be used to make predictions, meaning that it has been trained, optimized and evaluated on a specified test set.
\section{Implementation}
\label{sec:implementation}
\nirdi is implemented as a Web application that follows the architecture described in Section~\ref{sec:implementation}, so as to be a well structured, open, flexible and documented platform for \ppm supporting the capabilities described in Section~\ref{sec:capabilities}. 

We implemented the modules described in Section~\ref{sec:modules} in the back-end of our toolkit. Moreover, we provide a front-end web application that exploits the APIs made available by the toolkit to allow the user to interact with \nirdi through a state-of-the-art User Interface (UI).

The \emph{Front-end features} can be accessed via a Web browser that exposes the capabilities described in Section~\ref{sec:capabilities} through a graphical interface. The Web page is supported by React.js~\footnote{\url{https://reactjs.org}}.
The stateful component design of React.js allows us to load the interface and the state of the UI at two different times, allowing the user to browse the static content while waiting for the load of the dynamic content. After the dynamic content load, the UI renders the information requested by the user without any lag due to the state being stored locally, thus avoiding further requests to the Back-end. To support the continuous improvement of \nirdi we integrated Google analytics anonymized tracking of the user flow through the pages of the interface in order to identify the most common flows which can act as candidates for future improvements.

The \emph{Back-end modules} described in Section~\ref{sec:modules} are implemented in Python~\footnote{\url{https://python.org}}. 
We exploited the Django\footnote{\url{https://djangoproject.com}} framework to expose the APIs, synchronize the database with the single code modules, and create the elements of the architecture. 
The event logs are handled through the PM4py library~\cite{DBLP:journals/corr/abs-1905-06169}. The logs are then encoded and get ready for the predictive model training using the Pandas Dataframe objects\footnote{\url{https://pandas.pydata.org/pandas-docs/stable/reference/api/pandas.DataFrame.html}}. 
We exploit the Sklearn~\footnote{\url{https://scikit-learn.org}} init/fit/partial\_fit/predict common signatures to create, build, update, and evaluate the predictive models. In this way, enriching the available techniques with new Machine/Deep Learning algorithms, both included in Sklearn or taken from external libraries such as PyTorch or TensorFlow through a wrapper, is a simple procedure. 
The queue handling is done through the Redis~\footnote{\url{https://redis.io}} Library.
The database manager 
is Postgres~\footnote{\url{https://postgresql.org}}.

\section{Using \nirdi}
\label{sec:use_case}

In this section, we provide an overview of the tool interface. For the sake of presentation we focus on the features that characterize more \nirdi, that is model construction, comparative evaluation and prediction explanation. The reader interested in exploring all the \nirdi~features can do it either by assessing the tool at \url{http://research.nirdizati.org/} or by following the step-by-step tutorial at \url{https://bit.ly/tutorial_nirdizati}. 
 
\subsection{Model Construction}
\label{sec:use_case:train}
\begin{figure*}[tb]
    \centering
    \includegraphics[width=.85\linewidth]{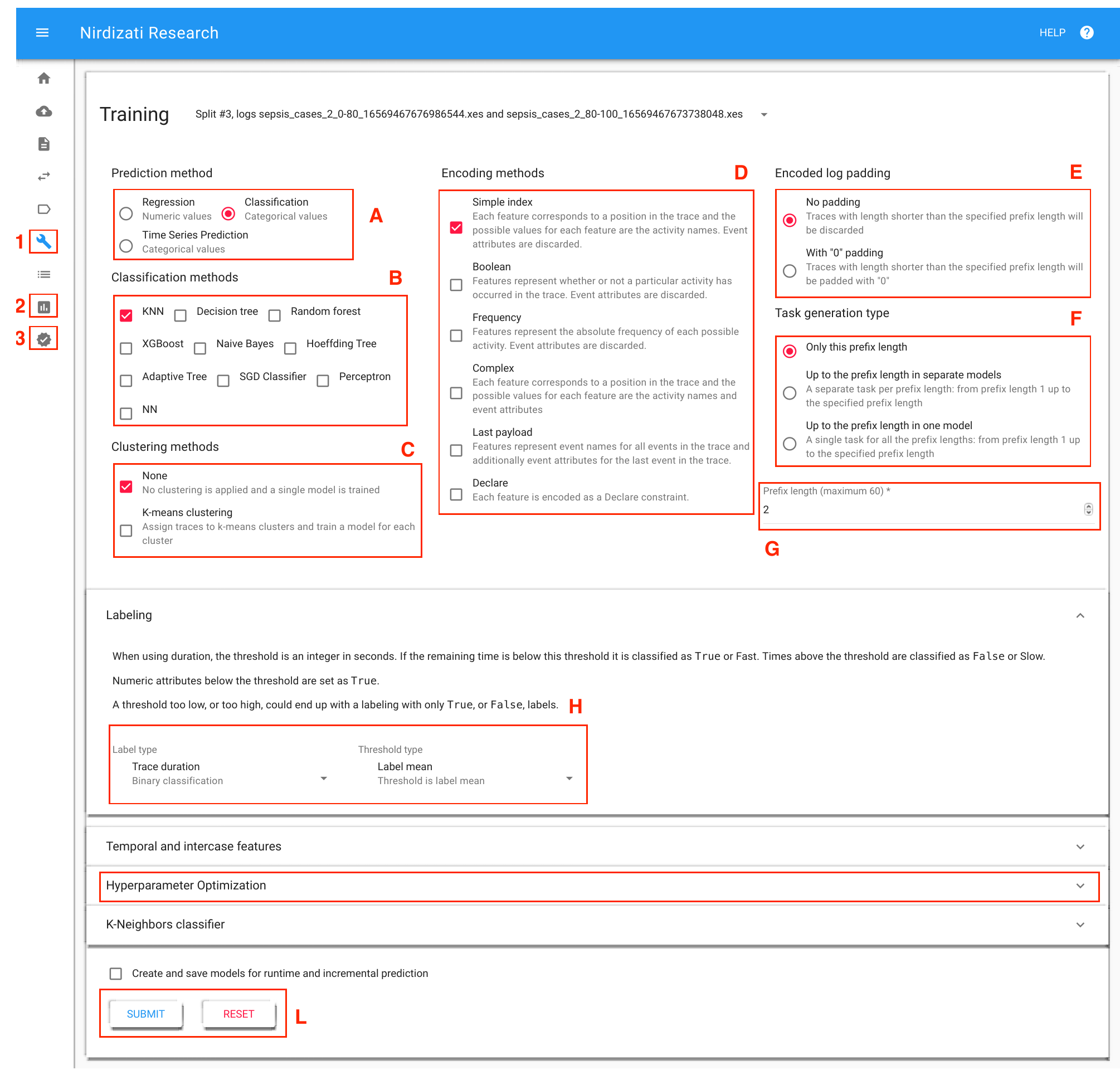}
    \caption{The \emph{Training page} of \nirdi.}
    \label{fig:use_case:classification1}
\end{figure*}

The \nirdi \emph{Training page}, accessible by clicking on the \textit{spanner} icon of the navigation menu (\rdb{1} in \figurename~\ref{fig:use_case:classification1}), enables the user to configure the construction of a predictive model. We illustrate the features provided in this page with the example of a predictive model for a binary configuration task with the help of~\figurename~\ref{fig:use_case:classification1}\footnote{We illustrate here the main features of the page leaving the rest to the tutorial at \url{https://bit.ly/tutorial_nirdizati}}. 

The \nirdi interface enables the user to select: the appropriate \textit{Prediction method} (\rdb{A}), the specific \textit{Classification method(s)} (\rdb{B}); whether to cluster the event log traces and, in this case, which \textit{Clustering method(s)} to use (\rdb{C}); the \textit{Encoding method(s)} (\rdb{D});  whether to use padding for the encoding (\rdb{E}); the \textit{Prefix length}(\rdb{G}); and how to deal with the prefixes up to the chosen prefix length (\rdb{F}). 
Finally, the interface enables the user to select the label to be used (\rdb{H}) in terms of \textit{Label type}, \textit{Attribute name}, and \textit{Threshold}, the type of hyperparameter optimization, here condensed for lack of space (\rdb{I}), and finally to \textsc{Submit} the request (\rdb{L}).

\subsection{Model Evaluation}
\label{sec:use_case:validate}
Clicking on the \textsl{histogram} icon of the navigation menu (\rdb{2} in \figurename~\ref{fig:use_case:classification1}), the user can open the \emph{Evaluation page}.
In this page the user can (filter and) select the predictive models to be evaluated and compared.

\begin{figure}[t!]
    \centering
    \includegraphics[width=.9\linewidth]{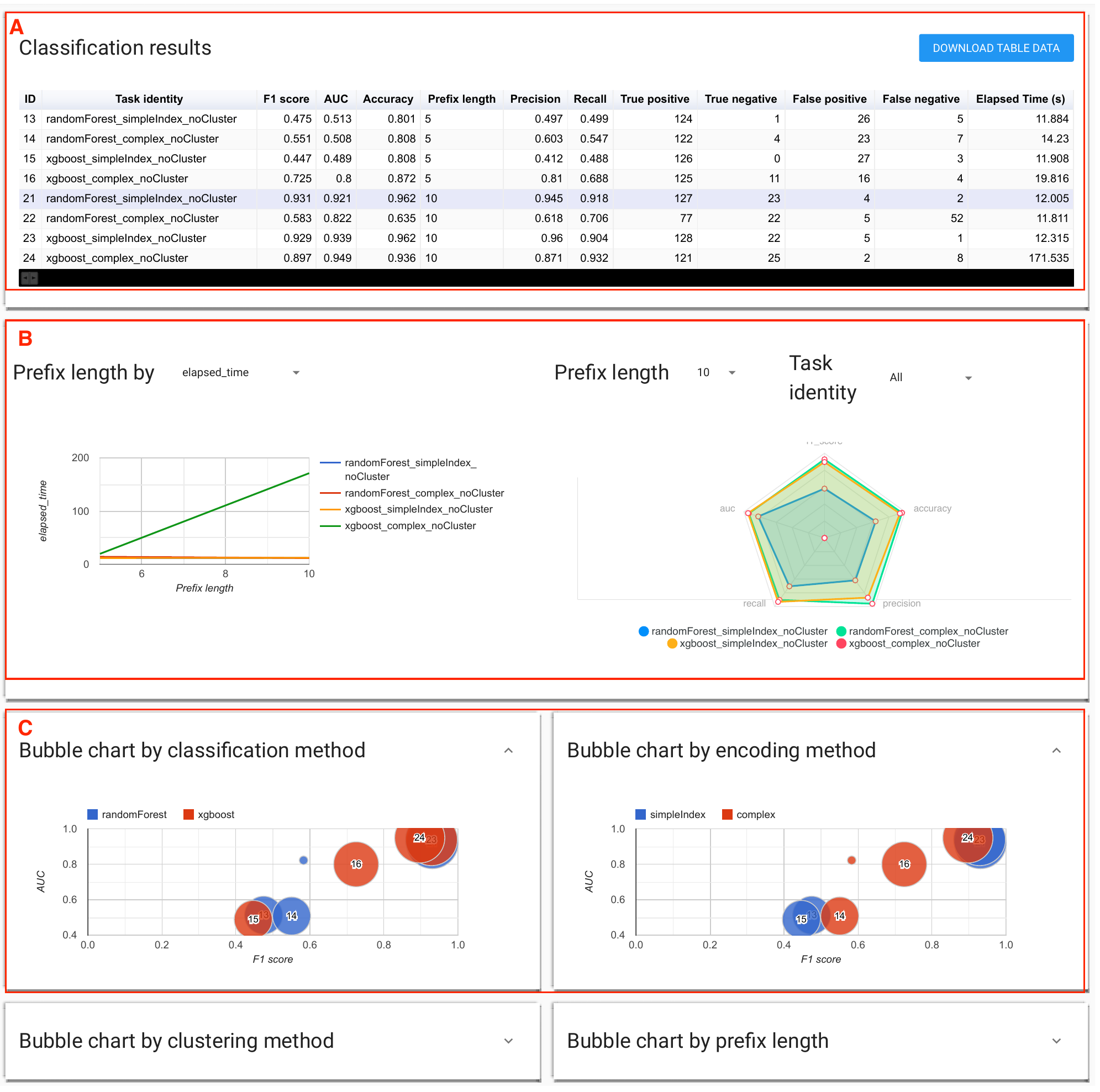}
    \caption{The features for the comparative evaluation of \nirdi.}
    \label{fig:use_case:validation3}
\end{figure}

For reasons of space we skip the model selection part of the page and focus on the comparative evaluation instruments, by assuming that the user is interested in comparing the models obtained by using 2 techniques (Random Forest and XGBoost), 2 encodings (simple and complex index), and 2 prefix lengths (5 and 10 events), thus resulting in $2^3$ predictive models. 
The names of the models in the table \emph{Classification results}, column \emph{Task Identity} (see \rdb{A}) in \figurename~\ref{fig:use_case:validation3} provide a hint of the configuration parameters selected for building the models.  

The results of the evaluation of the different models are reported in three different sections: the \emph{Result}, \emph{Prefix length} and \emph{Bubble Chart} sections. The \emph{Result} section (\rdb{A}) reports the values scored by the different predictive models on the test set, using a wide range of metrics, appropriate for the specific prediction task. This section allows the user to compare the metrics, to dynamically re-rank the models depending on the results in specific columns, and also to export them in a tabular format.
The \emph{Prefix length} section (\rdb{B}) displays the variation of performance depending on the prefix length. In particular the left-hand side diagram displays the trends of a specific metric (in \figurename~\ref{fig:use_case:validation3} the Elapsed time) for each predictive model across the chosen prefix lengths, while the right-hand side displays the performance of selected models (in \figurename~\ref{fig:use_case:validation3} all), for a specific prefix (in our case the four models having prefix length equal to 10). The larger the polygon, the better the model. Finally, the \textit{Bubble Chart} section~(\rdb{C}) displays four different bubble chart plots allowing the user to compare the models (identified by their Model ID) 
using different perspectives. In \figurename~\ref{fig:use_case:validation3} we can explore the comparison by focusing on the predictive model construction and encoding techniques used, respectively. The plots report on the two axes the AUC and F1 metrics, so that the closer to the top right corner, the better the model for both metrics.  

\subsection{Explanation}
\label{sec:use_case:explain}
\begin{figure}[t!]
    \centering
    \includegraphics[width=.9\linewidth]{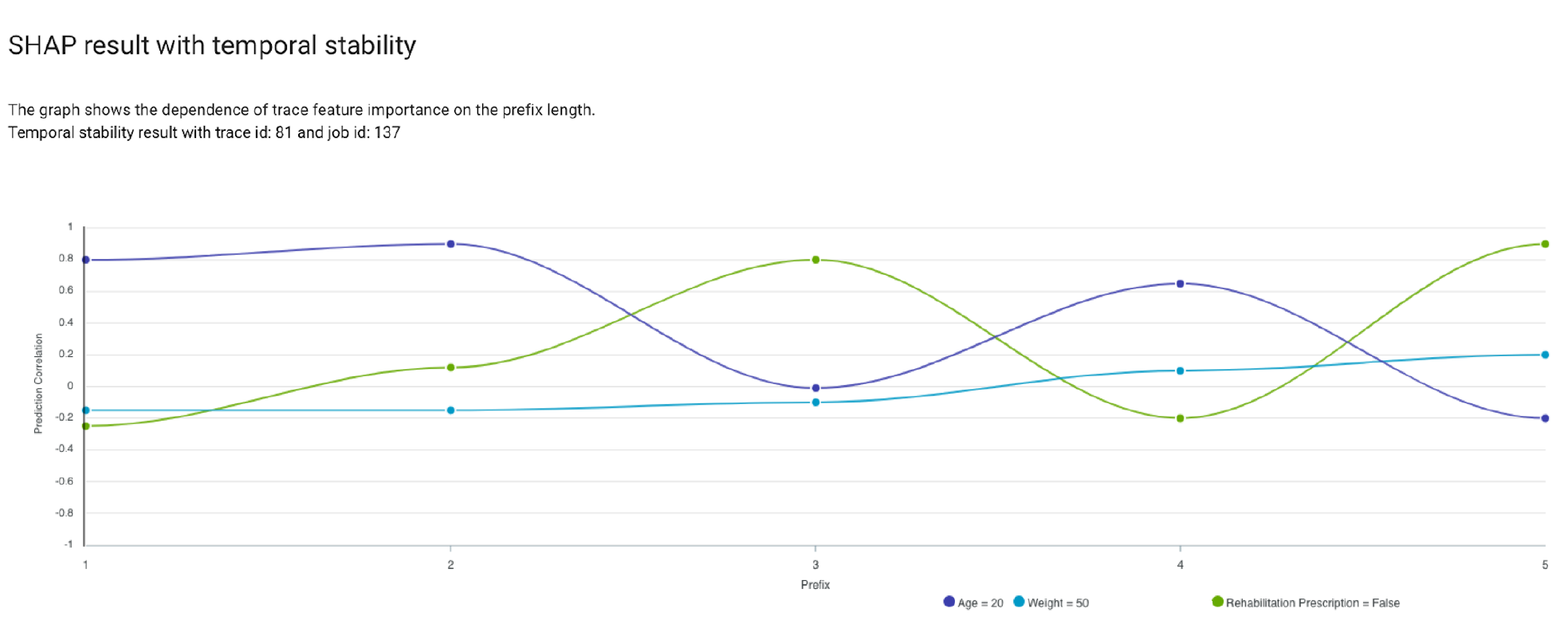}
    \caption{The trace level plot of the the \emph{Explanation page}.}
    \label{fig:use_case:explanation3} 
\end{figure}

Clicking on the \emph{verify} icon of the navigation menu (\rdb{3} in \figurename~\ref{fig:use_case:classification1}), the user can open the Explanation page from where she can select the predictive model used and the trace prediction to be explained (step omitted for space limitation). As illustrated in Section~\ref{sec:modules}, \nirdi supports explanations at the level of event, trace, and event log. \figurename~\ref{fig:use_case:explanation3} provides an example of the trace level explanation. 
The plot shows the correlation between 3 features (Age=20, Weight=50 and Rehabilitation Prescription=false) related to an incomplete trace of a patient who has carried out a treatment for a broken bone and the corresponding prediction (the patient will recover soon from the fracture) at different prefix lengths.
As we can see from the plot, at the beginning of the trace (prefix length $\leq 2$) the age of the patient positively correlates with the patient fast recovery, while the absence of a Rehabilitation Prescription has a negative impact on this outcome.
We can also note that the correlation of these two features with the prediction outcome vary in a considerable manner along the incomplete trace, while the trend of Weight=50 remains rather stable -- it does not seem to have a strong impact on the prediction along the whole prefix length. A detailed illustration of all the plots provided in \nirdi can be found in the aforementioned tutorial or in~\cite{rizzi2022explainable}. All the plots support the interaction with the user with hovering (to focus on an element) and clicking (to filter out unnecessary elements) behavior.  

\section{Assessment}
\label{sec:assesment}
The implementation of \nirdi is assessed through the use of reactive design patterns \cite{boner2014reactive} and load tests \cite{DBLP:journals/sttt/ShafiqueL15,DBLP:journals/sttt/SchieferdeckerDA05}. The reactive design patterns aim at assessing the interaction among the architectural elements, while the load tests are designed to emulate the work of multiple users accessing the prototype in a concurrent manner. 


\paragraph{Assessing the interaction among architectural elements}
\label{sec:implementation:elements_interaction}
The interaction among the architectural elements in \nirdi is not trivial, as illustrated in \figurename~\ref{fig:architecture:choreography}. 

We streamline the interaction by employing the reactive design patterns~\cite{boner2014reactive} that concern: 
\begin{enumerate*}[label=(\roman*)]
    \item fault tolerance/recovery, 
    \item control flow
    \item replication, 
    \item resource management, and 
    \item state management.
\end{enumerate*}

\textit{Fault Tolerance} is ensured by delegating the predictive models' training and testing 
to the Workers, which are completely separated from the rest of the toolkit. They are resilient towards code exceptions in the training/test of the predictive model and are instructed to save the experienced exceptions in the database. The logs can be monitored through a Grafana~\footnote{\url{https://grafana.com}} extension and, in case of unhandled exceptions, the \textit{Fault Recovery} is granted by the re-spawn feature of the Docker~\footnote{\url{https://docker.com}} daemon. The re-spawn of the Worker element and its connection to the rest of the toolkit is allowed by the Queue element, which can dynamically track the existence of any number of workers.

Concerning the \textit{Control Flow}, the Queue element in combination with the Worker, the Server, and the Database elements allows us to implement the \textbf{Pull} pattern to give the Worker the time to produce the predictive model. The Server element is not tied to wait for the Worker, which, once the training is finished, publishes its results in the Database element.

The capability to handle heavy workloads in terms of the amount of Jobs to carry out is ensured through the Master/Replica \textbf{Replication pattern}. The Master role is covered by the Server element. On the one side, it creates new jobs for the Workers and allows the workers to be aware of the existence of such jobs by saving the Job identifiers (IDs) on the Queue element; on the other side, it acts as an interface to the Front-end element exposing RESTful~\footnote{\url{https://restfulapi.net}} APIs. The Replica role is covered by the instances of the Worker element. The Worker element instances have two states, idle and running: when the Worker is in the idle state it is connected to the Queue element and waits for a new job ID to be published; when a new Job ID is available one Worker pops it from the Queue element, changes the Worker state into running and processes the Job.  
The details regarding the Jobs are stored in the Database element and are accessible through the Job ID inside the Back-end or through the Server element APIs. Such APIs can be accessed by the Front-end element which is a Web application allowing the user to access the Back-end in a facilitated manner. However, 
a more experienced user is not bound to use it. The experienced user can batch operations via simple scripts that can interact through the RESTful APIs with the Back-end without involving the Front-end.

Concerning \textit{Resource Management}, the event log and cached objects concurrent access is allowed by the Cache element. The Cache element is the sole owner of every resource in the toolkit, and the Cache element then implements the \textbf{Resource Loan} pattern allowing those who need the resource to access it but not to modify it. Also, the \textbf{Resource Pool} pattern is implemented: every time an element in the toolkit needs a resource it invokes the Cache element to access it. Indeed, there is no other way to access a resource. 

Finally, \textit{State Management} is ensured by the \textbf{Dissemination of Information} pattern which consists of storing multiple representations of the same object, which then allows for accessing such an object with a lower burden on the originating domain object (e.g., the original event log), thus enabling a faster and more flexible use of it. This pattern is applied on the event log objects: event logs are cached both in the pre-loaded and pre-processed form.

\paragraph{Assessing the workload}
\label{sec:implementation:performance_assessment}
\nirdi supports the users' requests by means of three different workflows: the upload workflow, the job computation request workflow, and the results browsing workflow.
The \emph{upload workflow} involves the Server element, the Database element, and the main memory of the host machine.
The \emph{job computation request workflow} involves the Server element and the Database element.
The \emph{results browsing workflow} involves the Server element, the Database element, and the Worker element. 

We did assess these three workflows by achieving a stable CPU usage above $80\%$ and an error per second $\le5\%$, as suggested in~\cite{DBLP:journals/sttt/ShafiqueL15} for stress and load testing. The stress test was performed using Vegeta\footnote{\url{https://github.com/tsenart/vegeta}}, while \nirdi was hosted on a machine equipped with 4 CPUs and 16GB of RAM. Table~\ref{table:stresstest} reports the results in terms of requests per second (rps) sustained by each workflow both at peak load and at sustained load. The tests show that the most computationally demanding workflow is the result browsing workflow. By assuming an average of 10-15 rps by user, we can conclude that \nirdi can support the concurrent usage of about 40 users.

\begin{table}[t]
    \centering
        \begin{tabular}{ l c c c c c c c c c c cc }
        \toprule
		Workflow & Peak Load & Sustained Load \\
        \midrule
        Upload             & 753 & 62 \\
        Job computation    & 767 & 64 \\
        Result browsing    & 750 & 60 \\
        \bottomrule
        \end{tabular}
    \caption{Assessing the workloads.}
    \label{table:stresstest}
\end{table}

\section{Related Work}
\label{sec:related_works}

The most popular tools that implement \PPM techniques are several ProM plug-ins, Apromore, IBM Process Mining Suite, and a Camunda plug-in.

ProM \cite{DBLP:conf/apn/DongenMVWA05} is a framework collecting a variety of independent plug-ins. Many of them are devoted to compute a specific \PPM prediction problem, that is, providing
\begin{enumerate*}[label=(\roman*)]
    \item numeric predictions \cite{DBLP:conf/ijcnn/PolatoSBL14, DBLP:conf/otm/FolinoGP12}; 
    \item outcome-based predictions \cite{DBLP:conf/caise/MaggiFDG14, DBLP:journals/ijcse/CastellanosSCDS06}; and
    \item next activity predictions \cite{DBLP:journals/computing/PolatoSBL18, DBLP:journals/is/LeoniAD16}.
\end{enumerate*}
Despite the aforementioned wide amount of plug-ins tackling single \PPM challenges, the only plug-in in ProM implementing multiple complex capabilities, such as the ones described in this work, is the one in \cite{DBLP:conf/bpm/FedericiRFDGMT15}. This plugin, however, only tackles the outcome and the numeric prediction problem, with no support for a scalable and modular architecture such as the one at the basis of \nirdi and no support for explanations.

Apromore \cite{DBLP:journals/eswa/RosaRADMDG11} is an advanced process mining tool that allows for the storage, analysis, and re-use of large sets of process models. 
The tool is web-based and therefore allows for the easy integration of new plug-ins in a service-oriented manner. 
This tool aims both at allowing practitioners to deal with the challenges of process stakeholders, and at enabling researchers to develop and benchmark their own techniques with a strong emphasis on the separation of concerns. 
The only plug-in performing \PPM related challenges in Apromore is the one in~\cite{DBLP:conf/caise/VerenichMRDRM18}. The plug-in performs outcome-based prediction, time-based prediction, next activity prediction, and the computation of performance metrics w.r.t. the event log. However, the amount of performance metrics available in this tool is limited; no help/indication on the pre-processing of the event log is given to the users, and there is no support for any explanation. Differently from \nirdi, Apromore also supports the model deployment phase. 

The IBM Process Mining Suite~\cite{galanti2021integration}~\footnote{\url{https://ibm.com/cloud/cloud-pak-for-business-automation/process-mining}} is a commercial software suite allowing for the discovery, conformance checking, monitoring, and simulation of processes. The suite is multilevel and provides business analysts with a digital twin of an organization (DTO). While this tool supports explanation and hyper-parameter tuning, differently from \nirdi, it mainly focus on a single predictive technique (CatBoost), thus lacking the capabilities related to model evaluation and comparison.

Finally, Camunda~\footnote{\url{https://camunda.com/}} is an open-source workflow and decision automation platform. The work in~\cite{DBLP:conf/caise/BartmannHCDD21} presents a plugin that allows for training, optimizing, and using \PPM models.
While the proposed plug-in is able of tackling all three types of predictions of \PPM, and automatically choosing the hyperparameters of the predictive model, it supports a limited amount of learning algorithms and a limited amount of evaluation metrics; moreover, no help/indication on the pre-processing of the event log is provided to the users and no support is given to explain the built predictive models.

\section{Conclusion}
\label{sec:conclusion}

This paper presented the latest advancements of \nirdi. Specifically, its Back-end has been redesigned to provide a reliable and efficient framework that allows for an easy comparison of different state-of-the-art Predictive Process Monitoring approaches and the inclusion of a wide set of \PPM techniques, by leveraging its modular architecture. 

The recent stream of publications in the \ppm~field~\cite{DBLP:conf/bpm/DiFrancescomarino18} shows the need for tools able to support researchers and users in analyzing, comparing and selecting the techniques that are the most suitable for them. This need is also reflected in the growth of \ppm~plug-ins in well known process mining tools such as ProM~\cite{DBLP:conf/apn/DongenMVWA05} and Apromore~\cite{DBLP:conf/caise/VerenichMRDRM18}.
\nirdi~is a completely dedicated tool for running a very rich set of \ppm~techniques and its latest advancements make it even more robust, scalable and usable.

We assess the current Technology Readiness Level of \nirdi~to be 5. 
This release offers indeed a well-defined structure of the software and code documentation;\footnote{\url{https://nirdizati-research.readthedocs.io/}} moreover, it is equipped with a very large test suite, and a \emph{Continuous Integration} deployment pipeline.
In addition to the workload tests presented in Section~\ref{sec:implementation:performance_assessment}, the tool has been extensively used and its features exercised on both simulated and real data~\footnote{The usage of \nirdi is monitored employing Google Analytics and the two years time span between the first prototype and the writing of this work recorded 14.637 sessions filtering bounce traffic and spending longer than a minute out of 23.871 sessions in total.} 
We believe that all these reasons make \nirdi~a mature and useful instrument for the BPM community.



\end{document}